\documentclass[10pt,twocolumn,letterpaper]{article}

\usepackage{3dv}
\usepackage{times}
\usepackage{epsfig}
\usepackage{graphicx}
\usepackage{amsmath}
\usepackage{amssymb}
\usepackage{multirow}

\usepackage[pagebackref=true,breaklinks=true,colorlinks,bookmarks=false]{hyperref}

\threedvfinalcopy 

\def\threedvPaperID{445} 

\ifthreedvfinal\fi
\begin{document}

\title{Road-aware Monocular Structure from Motion and Homography Estimation}

\author{Wei Sui$^{1}$, Teng Chen$^{1}$, Jiaxin Zhang$^{1}$, Jiao Lu$^{2}$,  Qian Zhang$^{1}$\\$^{1}$Horizon Robotics
$^{2}$China University of Geosciences\\{\tt\small
\{wei.sui, teng.chen, jiaxin02.zhang,
qian01.zhang\}@horizon.ai}, \tt\small{2004180013@cugb.edu.cn}
}

\maketitle

\begin{abstract}
   Structure from motion (SFM) and ground plane homography estimation are critical to autonomous driving and other robotics applications. Recently, much progress has been made in using deep neural networks for SFM and homography estimation respectively. However, directly applying existing methods for ground plane homography estimation may fail because the road is often a small part of the scene. 
  Besides, the performances of deep SFM approaches are still inferior to traditional methods. In this paper, we propose a method that learns to solve both problems in an end-to-end manner, improving performance on both. The proposed networks consist of a Depth-CNN, a Pose-CNN and a Ground-CNN. The Depth-CNN and Pose-CNN estimate dense depth map and ego-motion respectively, solving SFM, while the Pose-CNN and Ground-CNN followed by a homography layer solve the ground plane estimation problem. By enforcing coherency between SFM and homography estimation results, the whole network can be trained end to end using photometric loss and homography loss without any groundtruth except the road segmentation provided by an off-the-shelf segmenter. Comprehensive experiments are conducted on KITTI benchmark to demonstrate promising results compared with various state-of-the-art approaches. 
\end{abstract}

\section{Introduction}

Homography estimation is essential for inverse perspective mapping (IPM) \cite{jeong2016adaptive} and planar parallax estimation \cite{irani1996parallax} where a reference plane is referred, which further helps lane detection, bird's-eye view (BEV) generation and other vision tasks in autonomous driving.

Traditional homography estimation methods compute sparse feature correspondences between subsequent frames and apply the RANSAC algorithm ~\cite{fischler1981random}, which requires sufficiently large ego-motion. Furthermore, the effectiveness of such methods relies heavily on high quality feature correspondences, making it challenging in autonomous driving scenarios since the road surface is usually weakly-textured, occluded, or with repetitive patterns.

CNN-based methods treat homography estimation as a regression problem via the 4-point parameterization \cite{detone2016deep,nguyen2018unsupervised, zhang2019content}. By leveraging high level context features, these methods often outperform the traditional methods on challenging cases mentioned above. The drawback is that even though ego-motion and planar geometry can be recovered via homography matrix decomposition, such methods cannot provide true scale desired for autonomous driving.

Another branch of methods calculate homogrpahy matrix explicitly according to Eqn.~\ref{homography} which use ego-motion, ground plane and camera intrinsic parameters. While camera intrinsic parameters can be obtained via offline calibration, accurate ego-motion and ground plane estimation remains a challenge problem. 

To estimate ego-motion, Visual Odometry (VO) or Structure from Motion (SFM) are commonly utilized \cite{hartley2003multiple}. Sensors fusion with filter-based \cite{davison2007monoslam} or optimize-based frameworks \cite{klein2007parallel} can increase the accuracy and robustness. These methods usually requires large ego-motion as well. Recently, CNN-based methods have gained popularity. The supervised methods \cite{kendall2015posenet, li2017indoor, clark2017vidloc} regressed 6D pose directly, while the unsupervised methods \cite{monodepth2, godard2017unsupervised, li2018undeepvo} estimate pose and depth to construct supervisory signals via new view synthesis \cite{zhou2017unsupervised}. These methods can obtain accurate ego-motion even with small motion parallax, however they do not guarantee accurate homography estimation.

Similar to ego-motion estimation, ground plane also can be estimated from either geometry or learning based methods \cite{mcdaniel2010ground, se2002ground, man2019groundnet, xiong2020joint}. Geometry based methods obtain plane geometry by directly fitting a plane to the 3D points acquired from depth sensors or multi-view stereo. When only a single image is available, the ground normal (the plane coefficients) can be recovered by detecting vanishing points and the horizon line \cite{hartley2003multiple}. Recently, CNN-based methods are proposed to regress ground normal \cite{bansal2017pixelnet, qi2018geonet} from a single input image as well. Up to now, the performance of CNN-based methods are inferior to geometry based methods.

\begin{figure*}[t]
\begin{center}
   \includegraphics[width=1.0\linewidth]{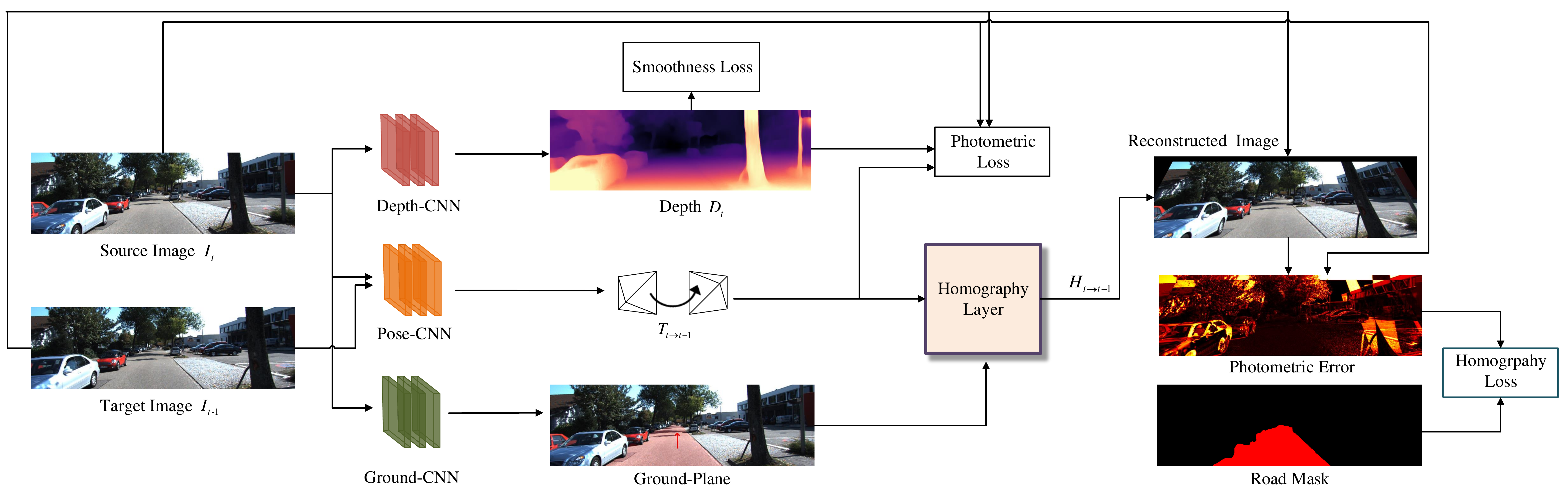}
\end{center}
   \caption{The overview of our network architecture. Our network consists of a Depth-CNN, a Pose-CNN and a Ground-CNN. The input are two consecutive images while the output are depth, ego-motion, ground plane as well as road homography. The output depth along with the ego-motion can be used to reconstruct the source image. While the homography and ego-motion are used to reconstruct the road surface. The road mask are provided by an off-the-shelf segmenter.}
\label{fig:network}
\end{figure*}

Our method also explicitly computes homography using ground plane and ego-motion. Differentiating from the aforementioned work, we propose to simultaneously optimize all these objectives in one network. In addition, we observe that adding depth estimation brings further improvement. The resulting multi-task network predicts depth, ego-motion, and ground plane at the same time. The depth and ego-motion sub-networks constitute learning-based SFM \cite{zhou2017unsupervised}, while the ego-motion and ground plane derives the ground homography matrix. The architecture of the network is demonstrated in Figure \ref{fig:network}. The entire network can be trained in a weakly-supervised manner, i.e., only requiring road segmentation from any off-the-shelf semantic segmention model to limit photometric loss to the road region, while completely eliminating the need for expensive 3D groundtruth.

Our main contributions are summarized as follows:
\begin{itemize}
    \item We propose a novel multi-task learning framework which can solve SFM (ego-motion and depth) and ground plane homography simultaneously. 
    \item We design the photometric loss, homography loss and depth-smoothness loss in a way such that the sub-networks provide supervisory signals for one another, hence the training can be done without any groundtruth other than a road segmenter.
    \item We show that joint learning can improve performance of all sub-tasks over existing methods through experiments on the KITTI benchmark dataset \cite{geiger2012we}.
\end{itemize}

\section{Related Work}
In this section, we introduce the works most relevant to our method including homography estimation, visual odometry and ground plane estimation. 

\subsection{Homography Estimation}
Homography describes the corresponding relationship of the points on the same plane imaged in two views. An increasing number of works focus on homography estimation to support other vision tasks, such as image stitching and IPM. We roughly classify existing methods into two categories: implicit methods and explicit methods.

Traditional implicit methods utilize geometry constraint to estimate homogrpahy. The most common idea is to first conduct keypoints detection and matching, and then find the optimal homography via RANSAC outlier rejection~\cite{fischler1981random}. This kind of methods is quite effective and robust when the features are of high quality. Nonetheless, the performance may degrade dramatically when the detected keypoints are insufficient or distribute unevenly, which is common phenomenon in driving scenes. Recently, inspired by the success of deep convolution neural network (CNN) in computer vision, deep homography estimation methods become prevalent \cite{detone2016deep,nguyen2018unsupervised, zhang2019content,le2020deep}. These methods directly regress the coordinates offset of specified 4 points~\cite{detone2016deep} according to the 4-points parameter estimation. However, a great improvement is demanded for the deep homography to be superior to traditional methods. On the one hand, implicit methods estimate homography directly from a image pair ignoring the true motion between two views. On the other hand the resulted homography may not be guaranteed corresponding to the road plane. The above two reasons make implicit methods unsuitable for applications where a reference plane is required.

Explicit methods \cite{hartley2003multiple,xiong2020joint} usually translate the homography estimation problem into two sub-problems: ego-motion and ground plane estimation, assuming the camera intrinsic parameters are known previously. Hartley et.al \cite{hartley2003multiple} first used the normal vectors of the coplanar points and ego-motion to construct a homography. Our method is also belong to the explicit methods, but we implement it via CNNs.

\subsection{Structure From Motion}

SFM has been a research hotspot for decades. Existing methods can be categorized into geometry-based and learning-based methods. Geometry-based methods optimize ego-motion by minimizing reprojection errors or photometric errors \cite{mur2015orb} between images. Some methods \cite{engel2017direct, engel2014lsd}achieve impressive performance with careful parameter tuning, but they are still prone to scale drift as well as scale ambiguity.

Recently deep learning based methods are popular for visual odometry estimation. The DeepVO method \cite{wang2017deepvo} is the first learning based method which utilizes a recurrent neural network to obtain ego-motion. Due to the use of temporal constraint, the results achieve good accuracy and smoothness. However, the dependence on labeled data hinders it's application to practice. To solve this problem, unsupervised method are proposed \cite{zhou2017unsupervised,monodepth2,xue2020toward}. The SFM-Learner proposed in \cite{zhou2017unsupervised} individually designed depth network and pose network and incorporated depth and ego-motion into loss function to generate supervisal signals. Chen et al. \cite{chen2019self} introduced optical flow into this framework aiming at reducing the impact of dynamic regions on ego-motion estimation. As the photometric loss over optical flow decrease much more easily than over depth and ego-motion, the backward gradient is dominated by optical flow. Gordon et al.~\cite{2019Depth} replaced optical flow with residual flow to strengthen the depth and ego-motion learning. Although these methods achieve a great success, they cannot recover the metric depth and ego-motion. 

Some methods \cite{sucar2017probabilistic,xue2020toward,roussel2019monocular}
also try to recover real scale of the scene by leveraging prior information such as height of the camera, etc. 

\subsection{Ground Plane Estimation}
Ground plane estimation methods can also be divided into geometry-based methods and learning-based methods. Geometry-based methods usually calculate the plane through a robust model fitting algorithm like RANSAC. McDaniel et al. \cite{mcdaniel2010ground} used the 3D point cloud from LIDAR to identify the ground plane. \cite{se2002ground} estimated the ground plane by using the 3D point cloud generated from stereo disparity. The accuracy of geometry-based method heavily depend on the quality of depth information.

Learning-based methods attempt to estimate the ground plane either directly or indirectly from other related tasks. \cite{haines2012detecting} learn a classifier to distinguish local planar image patches and their orientations. The Man et al. \cite{man2019groundnet} proposed the GroundNet to first combine convolutional neural network and geometric consistency about depth and normal for ground plane estimation. Similarly, our method also leverage a network to predict the ground plane.

\section{Method}

In this section, we firstly introduce the whole framework of our method including three networks and a novel homography layer. After that we give the explanation of the unsupervised training loss. Finally, we show that how our method can recover the real scale of the road scene.

Suppose there are two successive frames at time $t$ denoted as $I_{t-1}$ and $I_{t}$ respectively, the depth map corresponding to $I_{t}$ is denoted as $D_{t}$. The ego-motion between $I_{t-1}$ and $I_{t}$ is a rigid transform $\mathbf{T}_{t\to{t-1}}\in\mathbb{R}^{4\times 4}$. The ground plane in $t$-$th$ frame is parameterized by $\mathbf{N}_t$ and $h_t$, where $\mathbf{N}_t$ is a normal vector with two degrees of freedom. $h_t$ is the signed distance from the coordinate origin to the plane, which can be identical to the mounting height of camera.

\subsection{Network}
\textbf{CNN-SFM}  We use a Depth-CNN and a Pose-CNN similar to \cite{monodepth2}. The Pose-CNN takes two frames $I_{t-1}$ and $I_{t}$ as input and estimates the ego-motion $\mathbf{T}_{t\to{t-1}}$, while the Depth-CNN only takes one single frame $I_{t}$ as input and regresses depth map $D_t$. The rotation of $\mathbf{T}_{t\to{t-1}}$ is represented via angle-axis vector donated as $\alpha_t$, $\beta_t$, $\gamma_t$.

\textbf{Ground Plane Estimation Via Deep Correction} 
In our method, we adopt a learning-based manner to estimate the ground plane. Specifically, a Ground-CNN is designed to estimate the ground plane  $\mathbf{N}_t\in\mathbb{R}^{3\times1}$ and the mounting height of the camera $h_t$ to the ground plane. The ground plane in $t$-th frame can be represented as 
\begin{equation} \label{palne} 
    \mathbf{N}_t^T \mathbf{P} + h_t = 0, 
\end{equation}
where $\mathbf{P}\in\mathbb{R}^{3\times1}$ is the 3D points of the road surface in the $t$-$th$ frame.

Observing that the ground planes in driving scenes vary smoothly in most cases, we apply a residual block to learn the offset of ground plane parameters. Specifically, we estimate the variation of the normal and height of the ground plane instead of themselves since residuals are usually more easily to learn. This process can be formulated as 
\begin{equation} \label{normal-variation} 
    \mathbf{N}_t = \delta\mathbf{R}_t\Bar{\mathbf{N}} ,
\end{equation}
\begin{equation} \label{height-variation} 
    h_t = \delta h_t+\Bar{h} ,
\end{equation}
where $\Bar{\mathbf{N}}$ and $\Bar{h}$ can be obtained from initial calibration, and then keep unchanged. We ignore the roll component of $\delta\mathbf{R}_t$, since it would have no impact on the normal estimation.
\begin{equation} \label{height-variation} 
  \delta\mathbf{R}_t = \mathbf{R}(\delta r_t)\mathbf{R}(\delta s_t ),
\end{equation}
where $\delta r_t$, $\delta s_t$ and $\delta h_t$ are the outputs of the Ground-CNN.

The backbone of our Ground-CNN is ResNet-18 \cite{he2016deep}, which takes $I_t$ as input and outputs a 3-dim vector [$\delta r_t$, $\delta s_t$, $\delta h_t$] defined above. Compared to \cite{se2002ground} regressing the normal of each pixels, we simply regress the ground plane normal by leveraging the consistency between the normal and depth to increase the robustness.

\textbf{Homography Layer} We propose a simple but effective homography layer to regress homograpy matrix. The input of the layer are ego-motion ($\mathbf{R}_{t\to{t-1}}$ and $\mathbf{t}_{t\to{t-1}}$), ground plane ($\mathbf{N}_t$ and $h_t$) and the output is the homogrpahy matrix $\mathbf{H}_{t\to{t-1}}\in\mathbb{R}^{3\times3}$. 
According to the definition of homography with respect to ego-motion and a reference plane, we can get the road homography as follows:
\begin{equation}\label{homography}
    \mathbf{H}_{t\to{t-1}} = \mathbf{K}(\mathbf{R}_{t\to{t-1}} - \cfrac{\mathbf{t}_{t\to{t-1}}  \mathbf{N}_t^T}{h_t})\mathbf{K}^{-1},
\end{equation}
where $\mathbf{K}\in\mathbb{R}^{3\times3}$ is the intrinsic matrix of the camera, which is assumed to be calibrated in advance. The derivatives respect to ego-motion and ground plane in the  back-propagation can be derived as following (the subscripts are omitted for simplicity)
\begin{equation}
\begin{split}
    \cfrac{\partial(\mathbf{Hx})}{\partial \mathbf{R}} = -(\mathbf{Rx})^{\wedge},
\end{split}
\end{equation}

\begin{equation}
\begin{split}
    \cfrac{\partial(\mathbf{Hx})}{\partial \mathbf{t}}  =  - \cfrac{\mathbf{N x}}{h},
\end{split}
\end{equation}

\begin{equation}
\begin{split}
    \cfrac{\partial(\mathbf{Hx})}{\partial h} = \cfrac{\mathbf{KtN^TK}^{-1}\mathbf{x}}{h^2},
\end{split}
\end{equation}

\begin{equation}
\begin{split}
    \cfrac{\partial(\mathbf{Hx})}{\partial \mathbf{N}}  = -\cfrac{\mathbf{t x}}{h},
\end{split}
\end{equation}
where, $\mathbf{x}$ represents the homogeneous image coordinates of a pixel and $(\cdot)^{\wedge}$ means the corresponding skew matrix of a 3-dim vector. Note that there are no parameters to be updated in the homogrpahy layer. Since the homogrpahy layer is differentiable, in training phrase the derivatives will be propagated. In the inference phrase, the homography layer output the homography matrix.

\begin{table*}
    \begin{center}
        \begin{tabular}{|c|r|r|r|r|}
            \hline
            \multirow{2}*{Method} &
            \multicolumn{2}{|c|}{sequence 09} &
            \multicolumn{2}{|c|}{sequence 10} \\
            \cline{2-5} 
            & $ t_{err}(\%)$ & $ r_{err}(^{\circ}/100m)$ & $ t_{err}(\%)$ & $ r_{err}(^{\circ}/100m)$ \\ \hline
            ORB-VO  & 14.00 & 3.06 & 13.23 & 5.08 \\ \hline
            Zhou et al.\cite{zhou2017unsupervised} & 17.84 & 6.78 & 37.91 & 17.78\\
            Monodepth2\cite{monodepth2}  & 14.88 & 3.43 & 11.83 & 4.93  \\ 
            Ours & \textbf{5.11} & \textbf{2.12} & \textbf{5.19} & \textbf{2.62} \\
            \hline
        \end{tabular}
    \end{center}
    \caption{Evaluation results of visual odometry on KITTI test sequences. $t_{err}$ and $r_{err}$ represent Relative Translation Error (RTE) and Relative Rotation Error (RRE) respectively. The best results are reported in bold.}
    \label{tab:p_est}
\end{table*}

\subsection{Loss Function}
In our method, the framework is trained in an unsupervised manner. The training loss is composed of three components: photometric loss, homography loss and the scale loss.

\textbf{Photometric Loss}
The photometric loss is commonly used in unsupervised learning based visual odometry estimation methods \cite{monodepth2}. It measures the difference between target image and the reconstructed image from source image through the outputs of the network. Given ego-motion $\mathbf{R}_{t\to{t-1}}$ and the depth map $D_t$ of frame $I_t$, we can find a correspondence between two images through
\begin{equation}\label{eqn-reprojection}
    \mathbf{p}_{t-1} \backsim \mathbf{K}(\mathbf{R}_{t\to{t-1}}D_t(\mathbf{p}_t)\mathbf{K}^{-1}\mathbf{p}_t	+ \mathbf{t}_{t\to{t-1}}),
\end{equation}
where $\mathbf{p}_{t-1}$ and $\mathbf{p}_{t}$ are corresponding pixels in frames $I_{t-1}$ and $I_t$ respectively, and $D_t(\mathbf{p}_t)$ represents the depth of $\mathbf{p}_t$. Thus, frame $I_t$ can be reconstructed as 
\begin{equation}
    I_{t'}[\mathbf{p}_{t}] = I_{t-1} \langle \mathbf{p}_{t-1}	\rangle ,
\end{equation}
where $I_{t'}[\mathbf{p}_{t-1}]$ are color values at position $\mathbf{p}_{t-1}$, and $\langle \rangle$ is a bilinear sampling operator. 

Similar to \cite{zhao2016loss, godard2017unsupervised}, we also use a robust photometric error combining structural similarity (SSIM) \cite{wang2004image} and L1 norm between two images which is given by
\begin{equation} \label{photo-loss}
    E_p(I_t, I_{t'}) = \alpha \cfrac{1-SSIM(I_t, I_{t'})}{2} + (1 - \alpha) ||I_t - I_{t'}||, 
\end{equation}
where $\alpha$ is a hyper-parameter.

In addition, we rely on smooth depth assumption whereby we apply edge-aware regularization to the discontinuities in the predicted depth map $D_t$ by minimizing
\begin{equation}\label{smooth-loss}
    E_s(D_t) = |\partial_x d^*_t| e^{-\partial_x|I_t|}+|\partial_y d^*_t| e^{-\partial_y|I_t|}, 
\end{equation}
where ${\overline{d_t}}$ is the mean-normalized depth. $\partial_x(\cdot)$ and $\partial_y(\cdot)$ represent the gradient operator in horizontal and vertical direction respectively.

{\bf Homography Loss} Given an image $I_t$, we use a road map $M_t$ to remove the non-road region from the color image by
\begin{equation}\label{road-seg}
   \tilde{I}_t = M_t\odot I_t,
\end{equation}
where $M_t$ is a binary mask which can be obtained from a semantic segmentation model.

If two corresponding points $\mathbf{p}_{t-1}$ and $\mathbf{p}_{t}$ are located on the assumed road surface, they will satisfy
\begin{equation}
   \mathbf{p}_{t-1} = \mathbf{H}_{t\to{t-1}} \mathbf{p}_{t}.
\end{equation}
Similar to photometric error, we can reconstruct the road region by using
\begin{equation}\label{homo-warp}
   \tilde{I}_{t'}^h[\mathbf{p}_t] = \tilde{I}_{t-1}\langle\mathbf{p}_{t-1}\rangle,
\end{equation}
where $\tilde{I}_{t'}^h$ represents image warped by a homography.
Combining Eqn.\ref{road-seg} and Eqn.\ref{homo-warp} we get the homography loss as
\begin{equation}\label{homo-loss}
   E_h(I_t, I_{t'}) = |\tilde{I}_{t'}^h - \tilde{I}_{t}|,
\end{equation}
where $|\cdot|$ is the L1 loss.

\textbf{Scale Loss} Absolute scale of the scene is hard to estimate for monocular visual odometry. To solve this problem, we use the camera height $h_c$ to decide the absolute scale of the scene. $h_c$ can be reliably accquired when the road plane and camera center are obtained. To keep the scale between SFM and homography consistent, we dynamically adjust the scale of the estimated depth. We compute the relative scale between the calibrated $h_c$ and the predicted $h_t$ as follows
\begin{equation}\label{depth-scale}
    s = \cfrac{h_c}{h_t}, 
\end{equation}
Base on Eqn.~\ref{depth-scale}, the predicted depth can be adjusted by using
\begin{equation}
    \hat{D}_{t} = sD_t.
\end{equation}
Then, $\hat{D}_{t}$ is substituted to compute all the losses defined above. 

\textbf{Total loss} Combining Eqns.~\ref{photo-loss}, \ref{smooth-loss} and \ref{homo-loss}, the total loss between frame $I_{t-1}$ and $t_{t}$ is defined as
\begin{equation}
    E = \mu E_p + \lambda E_s + \xi E_h,
\end{equation}
where $\mu$, $\lambda$ and $\xi$ are balancing factors.

\begin{figure}[t]
\begin{center}
   \includegraphics[width=1.0\linewidth]{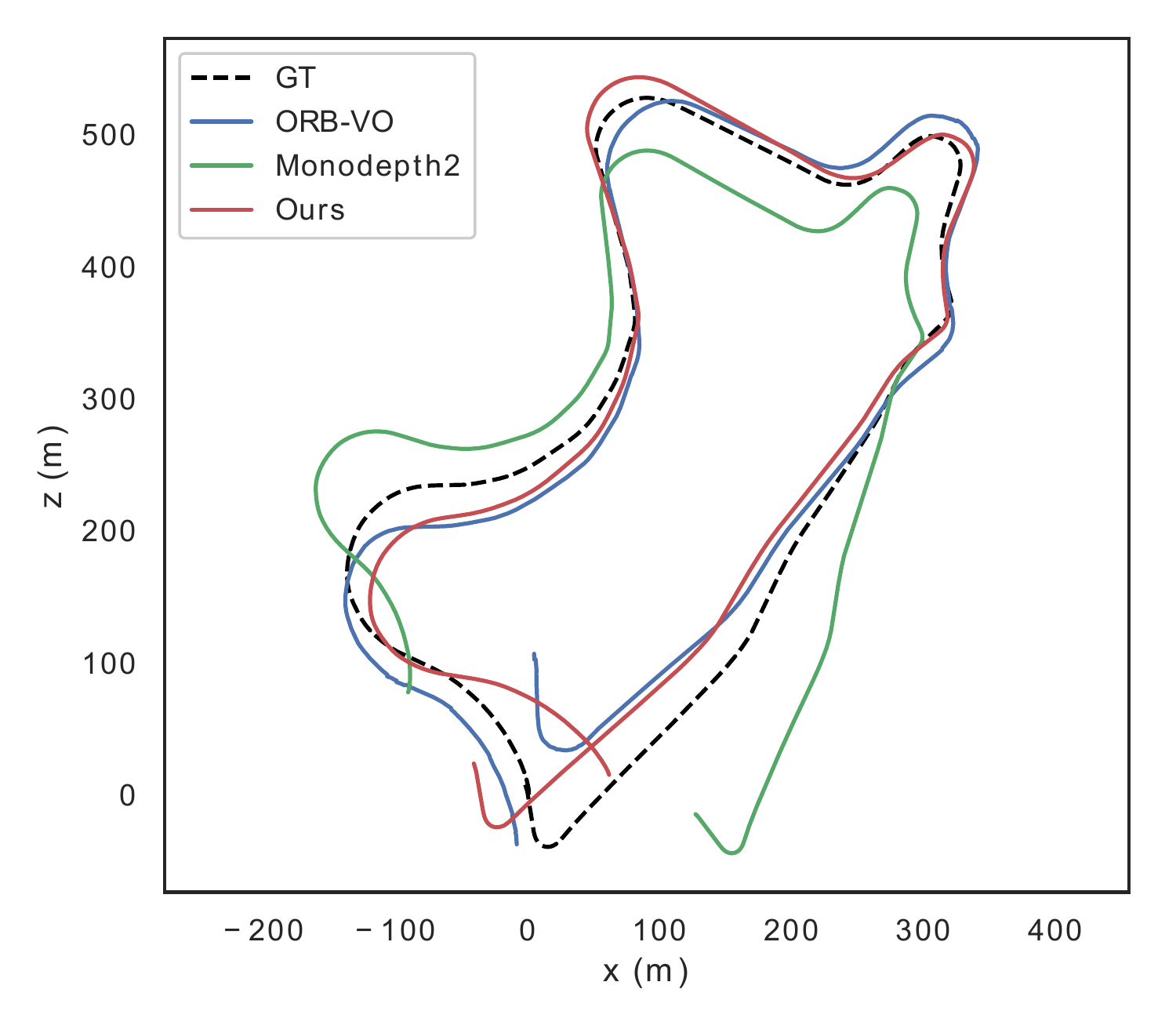}
   \includegraphics[width=1.0\linewidth]{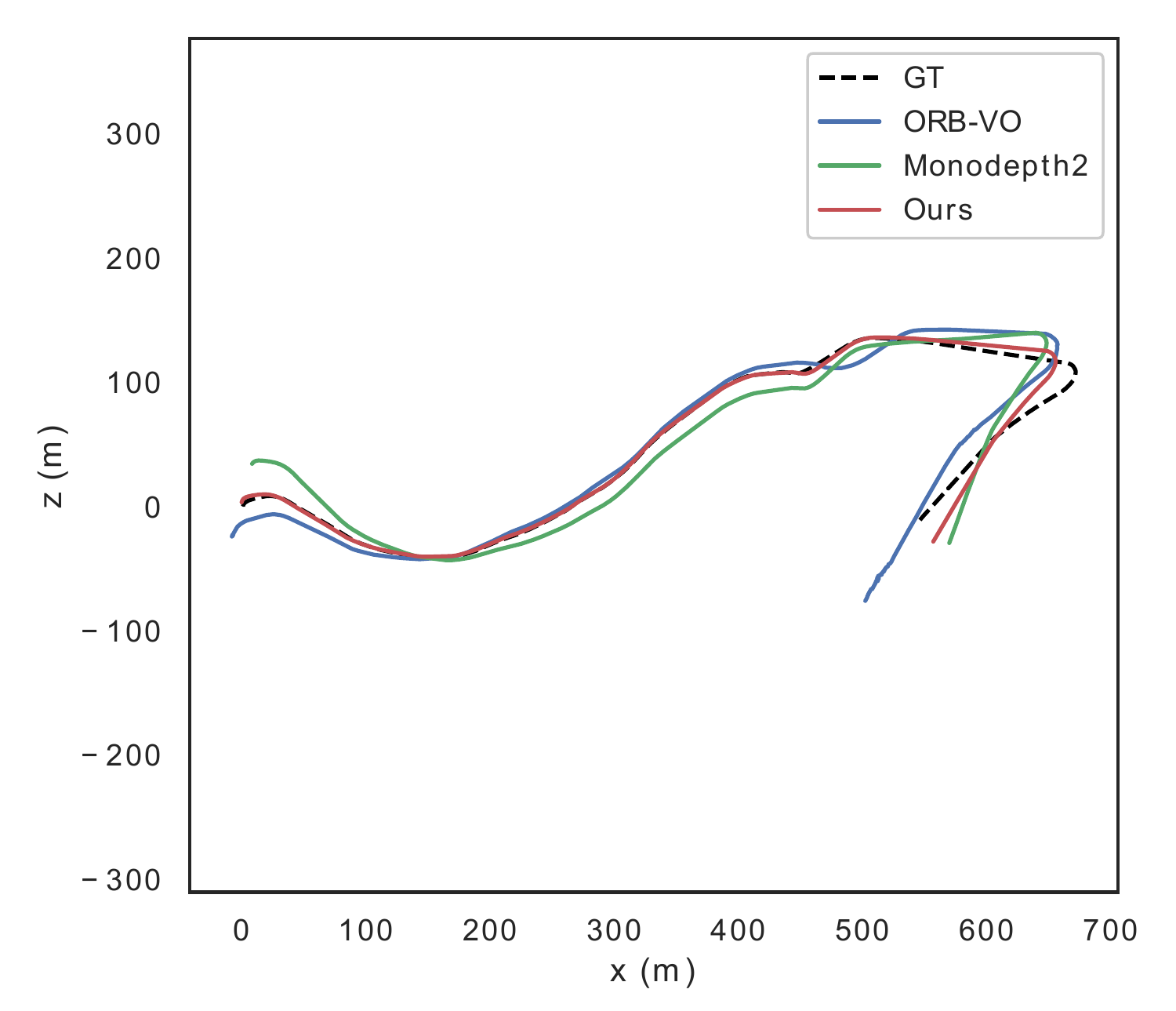}
\end{center}
   \caption{Trajectories of different methods on KITTI odometry test sequences. The top and bottom images show the results of sequence 09 and 10 respectively. Different colors represent the results of different methods}
\label{fig:pose}
\end{figure}

\begin{table*}
    \begin{center}
        \begin{tabular}{|c|c|r|r|r|r|r|r|r|}
            \hline
            \multirow{2}*{Method} &
            \multirow{2}*{Scale Factor} &
            \multicolumn{4}{|c|}{Lower is better} &
            \multicolumn{3}{|c|}{Higher is better} \\
            \cline{3-9} 
            && Abs Rel & Sq Rel & RMSE & RMSE log & $\delta < 1.25$ & $\delta < 1.25$ & $\delta < 1.25$ \\ \hline \hline
            Zhou et al.\cite{zhou2017unsupervised} &  GT &  0.183 & 1.595 & 6.709 & 0.270 & 0.734 & 0.902 & 0.959 \\ 
            Yang et al. \cite{yang2017unsupervised} & GT & 0.182 & 1.481 & 6.501 & 0.267 & 0.725 & 0.906 & 0.963\\ 
            Mahjourian et al.\cite{mahjourian2018unsupervised} & GT & 0.163 & 1.240 & 6.220 & 0.250 & 0.762 & 0.916 & 0.968\\ 
            DDVO \cite{wang2018learning} & GT & 0.151 & 1.257 & 5.583 & 0.228 & 0.810 & 0.936 & 0.974\\ 
            DF-Net \cite{zou2018df} & GT & 0.150 & 1.124 & 5.507 & 0.223 & 0.806 & 0.933 & 0.973\\ 
            GeoNet \cite{yin2018geonet} & GT & 0.149 & 1.060 & 5.567 & 0.226 & 0.796 & 0.935 & 0.975\\ 
            EPC++ \cite{luo2019every} & GT & 0.141 & 1.029 & 5.350 & 0.216 & 0.816 & 0.941 & 0.976\\ 
            Struct2Depth \cite{casser2019depth} & GT & 0.141 & 1.026 & 5.291 & 0.215 & 0.816 & 0.945 & 0.979\\ 
            CC  \cite{ranjan2019competitive} & GT & 0.139 & 1.032 & 5.199 & 0.213 & 0.827 & 0.943 & 0.977\\ 
            Bian et al. \cite{bian2019unsupervised} & GT & 0.128 & 1.047 & 5.234 & 0.208 & 0.846 & 0.947 & 0.976\\ 
            Monodepth2 \cite{monodepth2} & GT & 0.115 & 0.903 & 4.863 & 0.193 & 0.877 & 0.959 & 0.981\\ 
            Ours  & GT & \bf{0.111}  &   0.894  &   \bf{4.779}  &   \bf{0.189}  &   \bf{0.883}  &   \bf{0.960}  &   \bf{0.981}  \\
            \hline
            DNet\cite{xue2020toward}  & DGC & 0.118 & 0.925 & 4.918  & 0.199 & 0.862 &  0.953 & 0.979\\
            Ours  & H &0.128  &   0.936  &   5.063  &   0.214  &   0.847  &   0.951  &   0.978  \\
            \hline
        \end{tabular}
    \end{center}
    \caption{Depth evaluation results on the KITTI Eigen split \cite{eigen2014depth}. ``GT'' standards that results are usually reported using the per-image median ground truth scaling strategy introduced by ~\cite{zhou2017unsupervised}. ``DGC'' standards a method \cite{xue2020toward} to filter out the ground area and then calculate the median value of the camera height and ``H'' standards the predicted camera height by our method. The best results are reported in bold.}
    \label{tab:d_est}
\end{table*}

\section{Experiments}

We evaluate our methods from three aspects separately: depth and ego-motion (visual odometry) evaluation, homography, and ground normal evaluation. Comparative experiments with \cite{monodepth2} and \cite{mur2015orb} are conducted to show the effectiveness of our method. 

All the experiments are carried out on the KITTI benchmark which captured from various challenging scenes, e.g. highway, urban, and rural areas. Ground truth for visual odometry and depth evaluation are already provided. To the best of our knowledge, we are the first to evaluate homography and ground normal on KITTI dataset. 

To evaluate the homography estimation accuracy, we construct a homography validation dataset via feature based method, which is generated from the sequence 09 and 10 in KITTI's Odometry benchmark. Specifically, we first extract and match keypoints in the road region. After that we estimate the homogrpahy via 8-point algorithm and RANSAC algorithm \cite{fischler1981random}. To guarantee the reliability of the correspondence, we set the inlier threshold as 0.5 pixels. Finally, image pairs with inlier correspondences greater than 50 and keypoints evenly distributed will be selected to constitute the validation dataset. Similar to existing methods~\cite{ zhang2019content,le2020deep} the reprojection errors of the correspondences are utilized as a metric to evaluate the homography accuracy. 

We also generate ground-truth road plane for ground normal evaluation through a simple method. We first extract depth of the road region by leveraging the segmentation mask and then convert the depth to 3D points via inverse projection \cite{hartley2003multiple}. Following, we compute road plane by using the RANSAC \cite{fischler1981random} as well. Similar to homography, we discard images with a inlier ratio less than 0.6 which means the road surface is possible to be non-planar. The generated ground-truth road plane is believed to be reliable benifit from accuracy of the groundtruth depth.

\subsection{Implementation Details}
The whole framework is composed of three sub-networks: Depth-CNN, Pose-CNN, and Ground-CNN. The configuration of Depth-CNN and Pose-CNN has the same setting as in \cite{monodepth2}. The hyperparameters $\mu$, $\lambda$, $\xi$ and $\alpha$ are set to 1.0, 0.001, 0.1 and 0.85 in all our experiments. 
We train our model on NVIDIA GeForce GTX TITAN X for 20 epochs with batch size 8 and initial learning rate 0.0001. The Adam optimizer \cite{kingma2014adam} is used for training. 

\subsection{Depth and Visual Odometry Evaluation}
We follow KITTI benchmarks to perform quantitative depth and VO evaluation for our method. The proposed method is compared with both learning-based and traditional monocular methods. Among which Monodepth2 ~\cite{monodepth2} is the state-of-the-art learning-based method trained in unsupervised manner with monocular configuration, while the ORB-VO proposed in~\cite{mur2015orb} is the representative traditional VO method. It is worth noting that the backend and loop closure are removed from~\cite{mur2015orb} for the sake of fairness.

The VO evaluation results are reported in Table \ref{tab:p_est}. The peformance of ORB-VO surffer from serious degradation cause by the textureless, reflective or dynamic objects. In comparision, our method achieves the best results compared to traditional monocular methods and other unsupervised methods. From the comparative illustration in Figure.~\ref{fig:pose}, we can see that scale drift are apparently suppressed compared to ORB-VO and Monodepth2. This is because the ego-motion in our method is constrained from two aspects: homography constraint from the road region and the photometric from all the scene. Combining these two constraints help resist the negative affect on ego-motion estimation from the disturb of dynamic regions. 

Table \ref{tab:d_est} reports the comparative results between two groups of methods and ours. The first group is to recover the metric scale of estimated depth by using the per-image median ground truth scaling strategy denoted as "GT"~\cite{zhou2017unsupervised}. In this case, our method is trained without the prior knowledge of calibrated camera height. We can see that the accuracy of depth from our method are slightly better than that from Monodepth2~\cite{monodepth2} in most of the metrics. In second group, namely the last two rows in Table \ref{tab:d_est}, the estimated depth comes with metric scale from network. The DNet obtains the scale information by enforcing dense geometry constraints (DGC) \cite{xue2020toward}. Instead, our method recovers the scale during training process simply by using the ratio of the calibrated and estimated camera height. We can see that our method achieves comparable accuracy to DNet in a more concise way. 

\begin{figure*}
\begin{center}
    \includegraphics[width=1.0\linewidth]{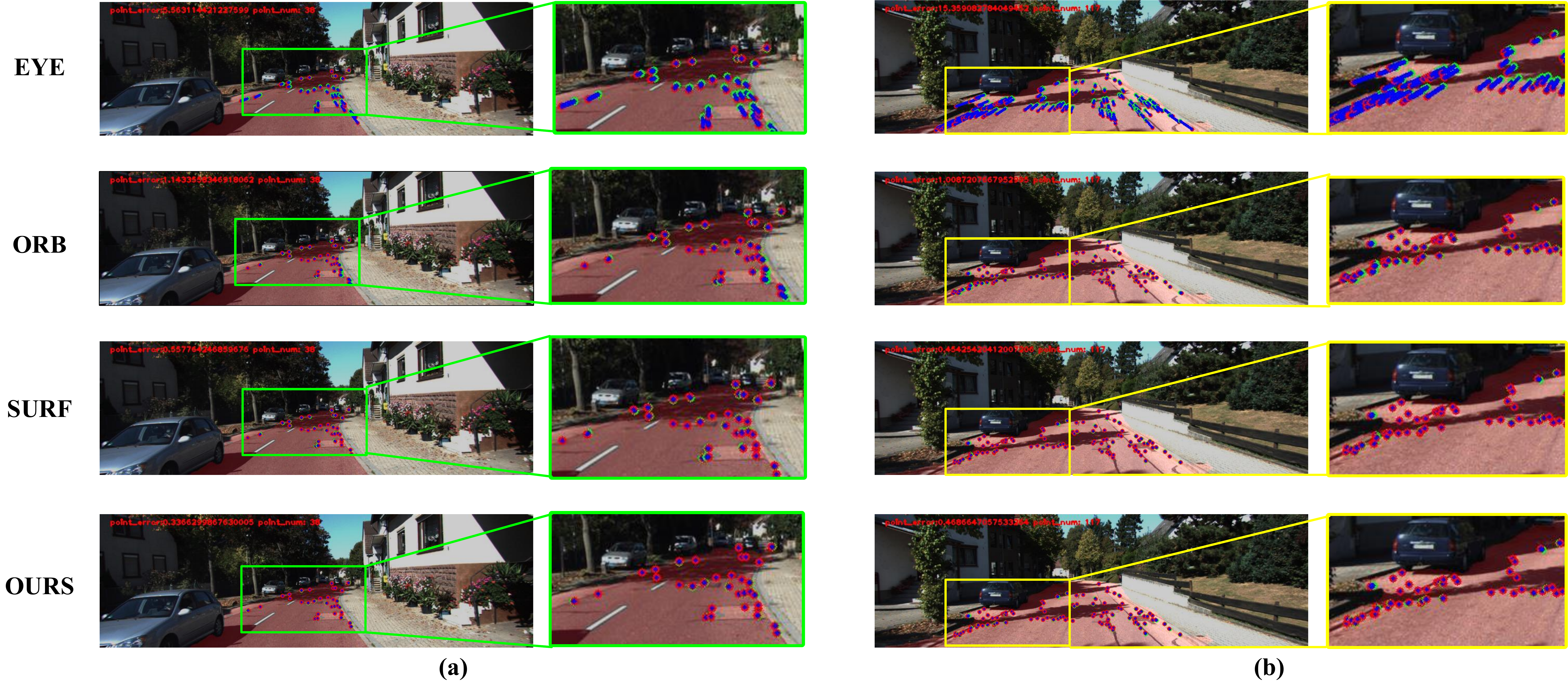}
\end{center}
   \caption{Examples of homography matrix.We drew source points (green) and  transformed target points (red), and connected them with a line (blue) on the picture. The length of the line indicates the estimation error. Colored rectangle depict areas for comparision}
\label{fig:homography}
\end{figure*}

\begin{figure*}
\begin{center}
    \includegraphics[width=1.0\linewidth]{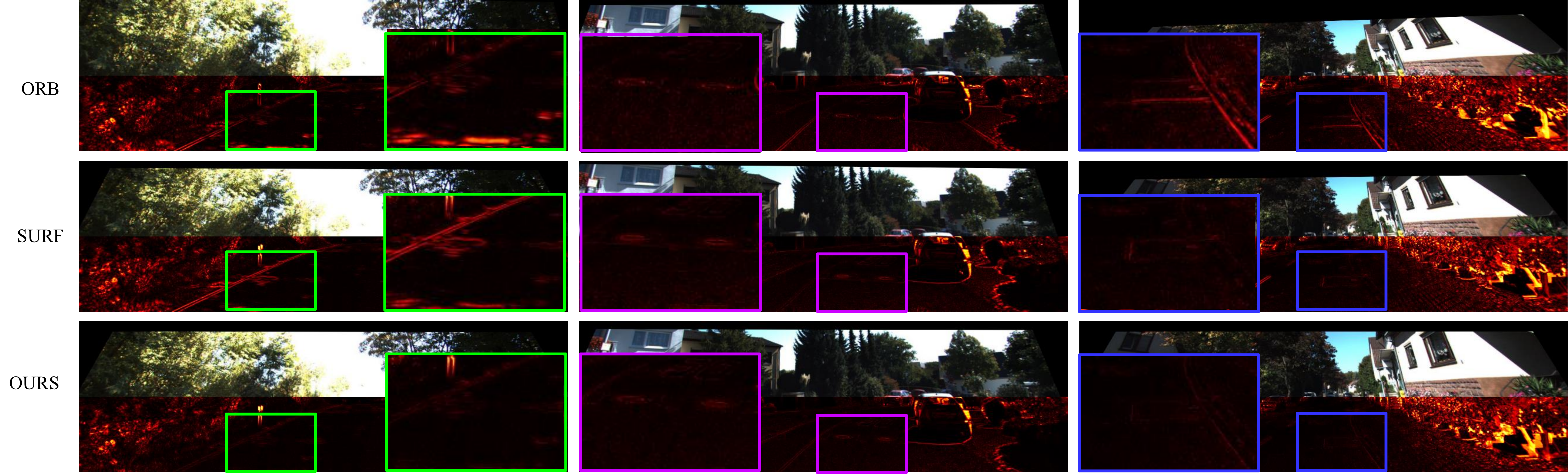}
\end{center}
   \caption{Visualization of homography photometric errors. We visualize the photometric error between the reconstructed image and the source image. The upper part of the picture is the reconstructed image, and the lower part is the visualized photometric error. From black to red indicates error from small to large. Colored rectangles depict areas for comparision.}
\label{fig:homography2}
\end{figure*}

\subsection{Homography Evaluation}

\begin{figure} 
\begin{center}
    \includegraphics[width=1.0\linewidth]{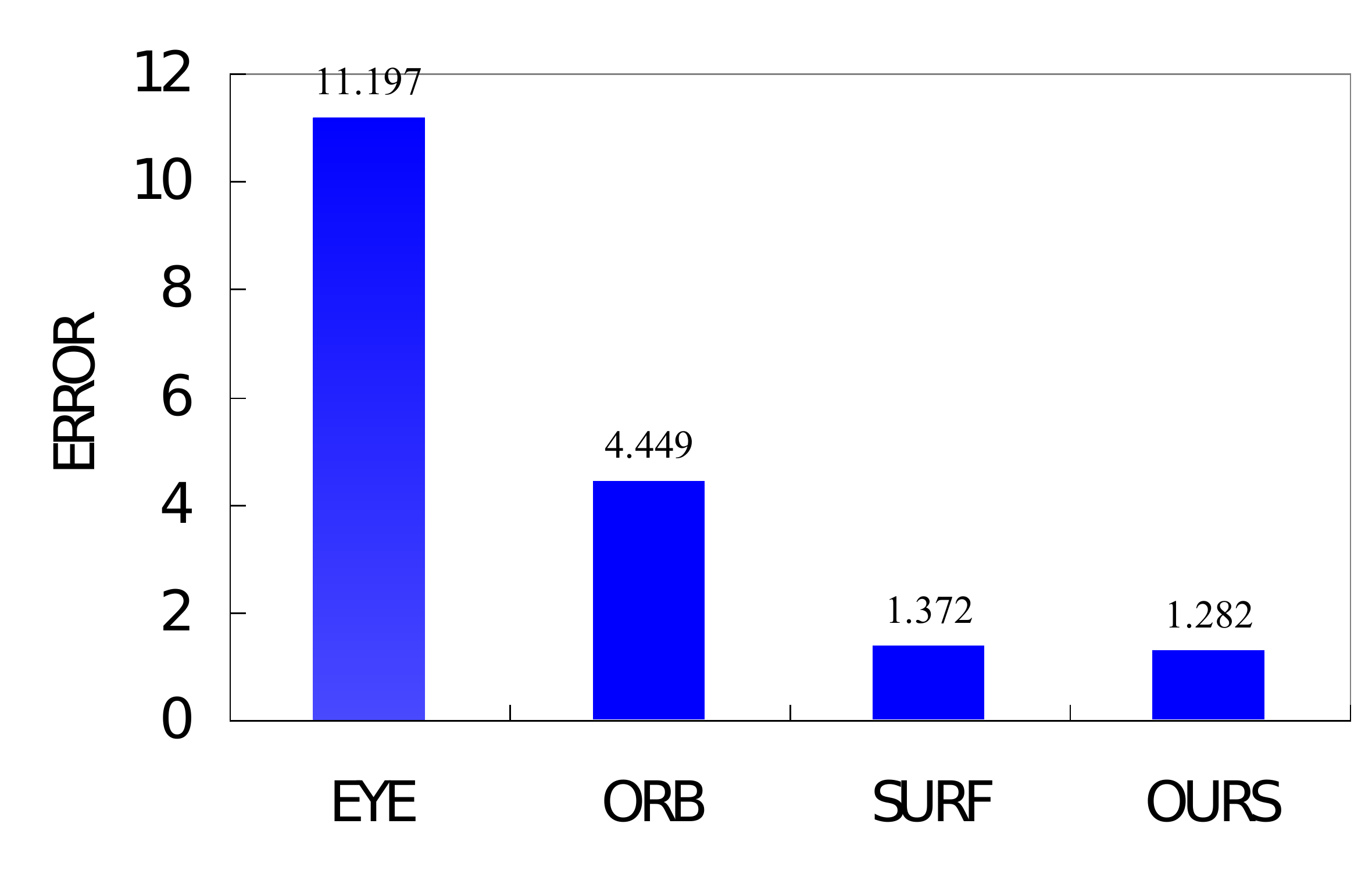}
\end{center}
   \caption{Keypoint reprojection errors of the homography estimation produced by exisiting methods and our method.}
\label{fig:homography_score}
\end{figure}

Since existing deep homogrpahy estimation methods are not aiming at a specific plane, they may fail when directly applied for road homography estimation because the road it not often dominant in the scene. Therefore we compare our methods with the traditional feature based methods using the reprojection error metric, which is depicted in
Figure.~\ref{fig:homography_score}. We can see that our method outperforms representative feature-based methods ORB and SURF, achieving the lowest reprojection error.

The visualized reprojection errors are compared in Figure.~\ref{fig:homography}. Observing from the enlarged local regions, we can see that the reprojected points almost completely overlap with the target points by using our method, while there are apparent offsets by using traditional methods. The reason might be that the extracted keypoints in feature-based methods distribute unevenly on the textureless road regions, which has a seriously negative impact on homography estimation. Instead, our method requires the estimated ego-motion and ground normal to meet densely road-aware homography constraint which leverages more effective information. The photometric error is also demonstrated in Figure.~\ref{fig:homography2}. It can be observed that the photometric errors near the road boundary area of our method are much smaller than feature-based methods, which further demonstrates the effectiveness of our road-aware homography estimation.

\begin{figure*}
\begin{center}
    \includegraphics[width=1.0\linewidth]{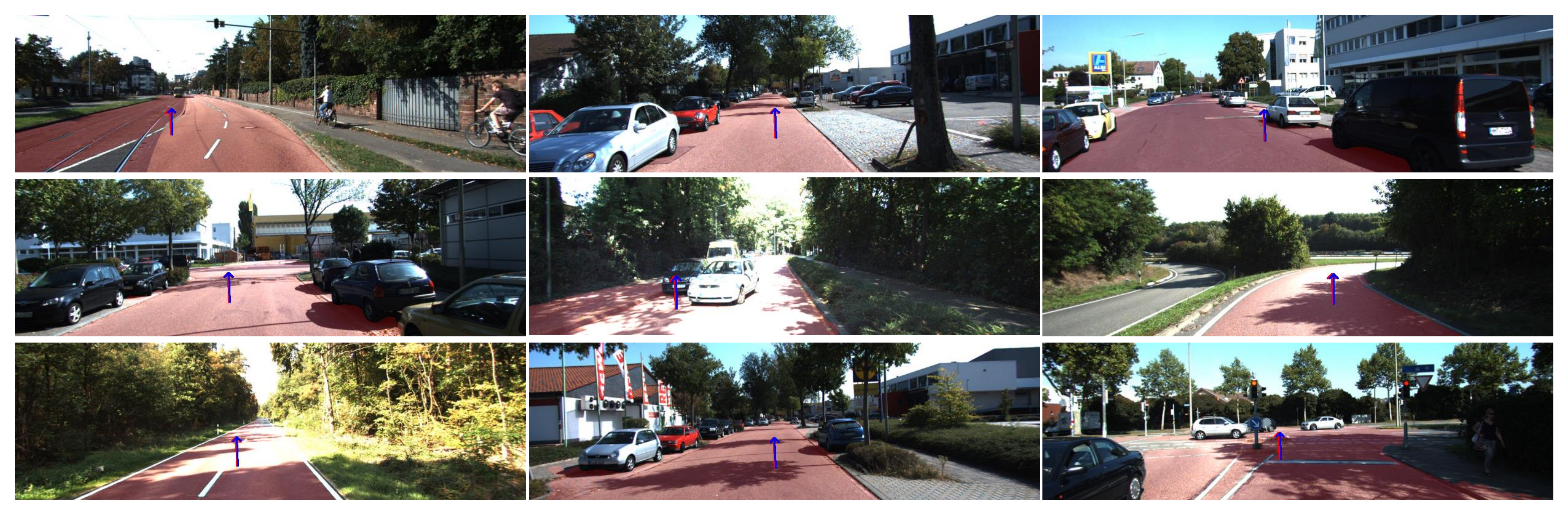}
\end{center}
   \caption{Illustration of estimated ground normal. The red arrow represents the predicted normal while the blue arrow represents the ground truth.
   For visual convenience, we shift the start point of predicted normals by 2 pixels to the right to avoid overlaps with ground truth labels.}
\label{fig:normal}
\end{figure*}

\begin{table}
    \begin{center}
        \begin{tabular}{|l|c|}
        \hline
        Method & Error/deg \\
        \hline\hline
        GroundNet\cite{man2019groundnet} (Supervised) & 0.70\\
        \hline
        HMM\cite{dragon2014ground} (Unsupervised) & 4.10 \\
        Lu Xiong et al.\cite{xiong2020joint} (Unsupervised) & 3.23 \\
        Ours (Unsupervised) & 1.12 \\
        \hline
        \end{tabular}
    \end{center}
    \caption{Evaluation result of ground plane estimation on KITTI dataset. The ground truth normal vector is calculated based on the ground truth external parameters}
    \label{tab:ground_est}
\end{table}

\subsection{Ground Plane Normal Evaluation} 
In this experiment, we verify the proposed method on test dataset used in GroundNet\cite{man2019groundnet}. We quantitatively compare our method with existing methods in Table \ref{tab:ground_est}. The angle between predicted and ground-truth normal is used as the evaluation metric. From Table.~\ref{tab:ground_est} we can see that the mean error of our method is 1.12 degree which are significantly smaller than other unsupervised methods~\cite{dragon2014ground} \cite{xiong2020joint}. Note that the performance of our self-supervised method is comparative to the supervised method GroundNet\cite{man2019groundnet}, which guarantee the accuracy. Though without groundtruth for training, the depth and homography estimation provide supervisory signal for Ground-CNN. The visualized results are shown in Figure \ref{fig:normal}, from which we can see that the predicted ground normal vectors are nearly identical to the ground truth even in the areas with significant varying or shadows (see the results in last row), which indicates the effectiveness and robustness of the proposed Ground-CNN.

\section{Conclusion}
In this paper, we propose a method that combines SFM and road homography estimation into a unified framework. Similar to \cite{zhou2017unsupervised}, the SFM part consists of a Depth-CNN and a Pose-CNN. Meanwhile, the Pose-CNN combine with a newly introduced Ground-CNN followed by a novel homography layer are used to solve the road homography. By designing a photometric loss, a dense homography loss and a depth smoothness loss, the whole network is trained end to end without groundtruth except for road segmentation provided by an off-the-shelf semantic segementation model (to limit the homography loss to the road pixels). In addition, the metric scale of depth and ego-motion can be recovered as long as the mounting height of camera is roughly calibrated. Extensive experiments are carried out on the KITTI benchmark to show that the proposed method significantly improves the accuracy of ego-motion and homography estimation. As future work, road or ground segmentation can also be jointly inferred by the network, making our approach fully unsupervised.  


{\small
\bibliographystyle{ieee_fullname}
\bibliography{paper}
}

\end{document}